\pdfoutput=1
\documentclass{article}

\usepackage[dblblindworkshop, final]{template/official/neurips_2026}
\workshoptitle{AI Foundations for Power Grids: From Models to Deployment at Scale}
\makeatletter
\renewcommand{\@noticestring}{Accepted at AI Foundations for Power Grids Workshop, NeurIPS 2026 (Oral presentation).}
\makeatother

\usepackage[utf8]{inputenc}
\usepackage[T1]{fontenc}
\usepackage{booktabs}
\usepackage{graphicx}
\usepackage{dblfloatfix}
\usepackage{microtype}
\usepackage{amsmath,amssymb}
\usepackage{hyperref}
\usepackage{xcolor}

\newcommand{\bench}{FLARE}

\title{FLARE: Physics-Grounded Evaluation of Language-Model Agents for Feeder Restoration with Degraded Observations}
\author{%
  BHVSP Subrahmanyam \\
  MU-VT Interdisciplinary Advanced Research Centre for Transformative Technologies \\
  Mahindra University, India \\
  Indian Institute of Technology Madras, India \\
  \texttt{21f3000492@ds.study.iitm.ac.in}
  \And
  Neeraj Choudhary \\
  Department of Computer Science and Engineering \\
  Mahindra University, India \\
  \texttt{neeraj.choudhary@mahindrauniversity.edu.in}
}

\begin{document}
\maketitle

\begin{abstract}
Restoring service after distribution-grid faults requires decisions from incomplete and conflicting information while respecting electrical constraints. Addressing this need, we introduce \bench{} (Feeder-Level Agent Restoration Evaluation), a physics-grounded benchmark for post-isolation feeder restoration. \bench{} contains 800 AC-feasible cases across five observation settings and allows agents to gather evidence, evaluate switches, and perform restoration actions under fixed operational limits. We evaluate five instruction-tuned models over 78,750 episodes, including topology, distributed-generation, and loading shifts. Mean normalized regret ranges from 0.082 to 0.148. Although every model has zero median regret, P95 regret ranges from 0.814 to 1.000, revealing an important failure tail. Shadow replay shows that 96.4\% of 15,706 guard-rejected actions would have caused a physical violation. \bench{} enables reproducible evaluation of language-model-based restoration support while emphasizing the need for external safety checks. Code is available at \url{https://github.com/dr-pandit-69/FLARE-neurips}.
\end{abstract}

\section{Introduction}

Smart grids involve many connected tasks including forecasting, monitoring, fault management and service restoration. During a disturbance, these tasks come together in the operator workflow. Operators must combine evidence from monitoring systems and field reports, determine whether more information is needed and select actions without violating electrical limits. Language-model agents could support this process by interpreting different sources of information, using operational tools and proposing actions over multiple steps. Evaluating agents for this role therefore requires examining not only the final restoration outcome but also how they gather information, make sequential decisions and respond to safety constraints.

Existing research provides two important foundations for this problem. FLISR studies have developed optimization methods, expert-designed rules and learning-based approaches for identifying faults and computing restoration plans and switching sequences~\citep{zidan2017flisr,poudel2020restoration}. Power-grid benchmarks such as L2RPN evaluate sequential topology control while general language-agent benchmarks such as AgentBench test planning and tool use in interactive environments~\citep{marot2020l2rpn,liu2024agentbench}. More recent benchmarks evaluate language agents on live energy data, executable power-engineering problems and power-system dynamic studies~\citep{curcio2026energyagentbench,trashchenkov2026powersystems,zhang2026poweragentbenchdyn}. Across these lines of work, a key evaluation need remains: connecting the information available during restoration with an agent's sequence of decisions and the physical consequences of those decisions. We focus on the post-isolation stage of FLISR where an agent must decide what additional evidence to request and which switches to operate using SCADA measurements, customer calls and fault indicators that may be incomplete, delayed or conflicting.

To address this need, we introduce \bench{} (Feeder-Level Agent Restoration Evaluation), a physics-grounded benchmark aligned with the operator workflow for post-isolation distribution-grid restoration. Our contributions are as follows:
\begin{enumerate}
  \item[(i)] We create a dataset of restoration cases that combines diverse grid conditions with multiple observation settings, allowing agents to be evaluated when operational evidence is complete, limited or conflicting.
  \item[(ii)] We provide an interactive environment in which agents can gather evidence and issue sequential switching actions, with tie closures screened by nonlinear AC power flow and topology and fault-isolation guards.
  \item[(iii)] We define evaluation measures that separate the cost caused by limited observations from the additional cost introduced by agent decisions, allowing restoration quality, information use and safety violations to be analyzed separately.
\end{enumerate}

\section{The \bench{} Dataset and Restoration Task}

\paragraph{Dataset and case construction.}
Each physical case is constructed by pairing a switchable feeder operating point with one or more faulted sections and simulating protection-assisted isolation. The resulting record contains the post-isolation topology, affected customers, available switching actions, operation times and the load and generation state used by the AC simulator. Power-flow-validated restoration references are computed for each case before observation variants and sequential agent episodes are generated.

Measured German medium-voltage time series set the operating conditions on CIGRE and SimBench networks~\citep{thurner2018pandapower,meinecke2020simbench}; faults, protection actions, and switching outcomes are simulated. The 800-case primary set spans 293 feeder-years and contains 465 equipment, 262 third-party, and 73 weather events, with concurrent faults in 10.9\% of cases. Every case is unique and AC-feasible, with a nonempty feasible restoration reference; Appendix~\ref{app:evaluation} defines when the search certifies optimality.

\paragraph{Observation settings and evaluation sets.}
Each \emph{physical case} fixes the network, fault, post-isolation topology, available switches, and oracle reference. Settings alter only reports: clean; sparse SCADA; noisy channels; stale SCADA-age tags; or lower coverage plus higher errors (conflicting). The physical state and reference stay fixed across settings for paired comparisons. An \emph{episode} is one controller trajectory on a case and setting; Appendix~\ref{app:data} details the report protocol.

An \emph{evaluation set} is a collection of physical cases selected to answer a particular robustness question. The 800-case primary set is evaluated under all five observation settings and supports the main paired model comparison. The 150-case independent set contains separate fault events and checks whether conclusions extend beyond the primary events. The three 200-case operating-shift sets alter one factor at a time after the benchmark is frozen: normal-open topology, DER-to-load ratio or loading level. The four 100-case network-type sets test transfer across different feeder structures. Interface diagnostics use the primary cases while changing the prompt, memory, candidate ranking or tool availability, and stochastic sensitivity repeats 200 cases using sampled decoding. The counts in Table~\ref{tab:sets} refer to physical cases rather than episodes. The number of episodes is larger because a physical case can be evaluated across multiple observation settings, models and controller configurations. Exact construction, filtering and feasibility procedures are provided in the appendix.

\paragraph{Restoration task and agent interface.}
After fault isolation, the agent gathers reports, inspects or tests locations, simulates and operates listed switches under a 240-minute horizon and fixed budgets. Sensing, travel, switching, and rejected attempts consume time. Candidate actions are ranked by reported de-energized areas; the fault location and full physical state remain hidden except in diagnostics.

\paragraph{Scoring and safety.}
\bench{} measures class-weighted controllable energy not served over 240 minutes, using 1.2~kW per customer and residential/mixed/commercial/industrial weights of 1/1.5/2/3. Let $C_\pi$ be episode cost, $C^*$ the lowest feasible oracle-plan cost found for that physical case (shared across settings), and $C^0$ the no-action cost. Appendix~\ref{app:evaluation} gives the search and its optimality certificate. Normalized regret is
\begin{equation}
R(\pi)=\operatorname{clip}_{[0,1]}\frac{C_\pi-C^*}{C^0-C^*}.
\end{equation}
where zero matches the oracle reference and one is no better than taking no action. Greedy, random-feasible, FLISR, and belief-planner references use the same timing and score. A partial-information reference selects the best of acting on the most likely fault, sequential inspection, and switch testing; it is certified only within this class, covering 465/800 primary cases per setting (58.1\%).

Before a tie closure commits, the guard checks radiality and fault energization, then AC power flow on a trial copy of the simulated physical network: convergence, line/transformer loading at most 100.5\%, and energized-bus voltage in $0.90$--$1.10$~pu. This check is independent of agent reports. Sectionalizer openings instead follow dark-cut rules; rejected attempts leave the state unchanged and consume time.

\begin{figure*}[!b]
  \centering
  \begingroup
  \definecolor{flareblue}{HTML}{3B528B}
  \newcommand{\heat}[2]{\colorbox{flareblue!#1}{\makebox[1.05cm][c]{\scriptsize\strut #2}}}
  \newcommand{\heatw}[2]{\colorbox{flareblue!#1}{\makebox[1.05cm][c]{\scriptsize\strut\textcolor{white}{#2}}}}
  \setlength{\tabcolsep}{2.2pt}
  \renewcommand{\arraystretch}{1.05}
  \scriptsize
  \begin{tabular}{@{}lcccccccc@{}}
    \toprule
    & \multicolumn{5}{c}{Primary observation setting} & \multicolumn{3}{c}{Operating shift} \\
    \cmidrule(lr){2-6}\cmidrule(l){7-9}
    Model & Clean & Sparse & Noisy & Stale & Conflict & Topology & DER ratio & Loading \\
    \midrule
    Llama 3.1-8B & \heat{33}{0.065} & \heatw{66}{0.131} & \heatw{69}{0.138} & \heatw{65}{0.129} & \heatw{79}{0.158} & \heatw{70}{0.140} & \heat{54}{0.108} & \heatw{71}{0.141} \\
    Mistral Small 24B & \heat{30}{0.060} & \heat{46}{0.092} & \heat{47}{0.093} & \heat{48}{0.096} & \heat{49}{0.098} & \heat{49}{0.098} & \heat{39}{0.078} & \heat{53}{0.105} \\
    Qwen 2.5-3B & \heat{51}{0.102} & \heatw{70}{0.139} & \heatw{81}{0.161} & \heatw{82}{0.164} & \heatw{86}{0.172} & \heatw{77}{0.153} & \heatw{63}{0.125} & \heatw{80}{0.160} \\
    Qwen 2.5-7B & \heat{30}{0.060} & \heat{42}{0.084} & \heat{45}{0.089} & \heat{44}{0.088} & \heat{45}{0.089} & \heat{45}{0.090} & \heat{36}{0.071} & \heat{49}{0.097} \\
    Qwen 2.5-14B & \heat{33}{0.066} & \heat{58}{0.116} & \heat{54}{0.108} & \heatw{62}{0.124} & \heatw{70}{0.140} & \heatw{63}{0.126} & \heat{48}{0.095} & \heatw{63}{0.125} \\
    \bottomrule
  \end{tabular}
  \endgroup
  \caption{Mean normalized regret (lower is better; common 0--0.20 color scale). Primary columns use 800 cases; shifts average five settings over 200 cases. Color does not denote significance.}
  \label{fig:heatmap}
\end{figure*}

\begin{table}[!b]
  \centering
  \small
  \caption{FLARE evaluation sets.}
  \label{tab:sets}
  \renewcommand{\arraystretch}{1.05}
  \begin{tabular}{@{}l r p{3.4cm}@{}}
    \toprule
    Evaluation & Cases & Purpose \\
    \midrule
    Primary & 800 & Five-setting model comparison \\
    Independent & 150 & Separate fault events \\
    Operating shifts & $3\!\times\!200$ & Topology, DER ratio, and loading \\
    Network types & $4\!\times\!100$ & Four feeder structures \\
    Interface diagnostics & 800 & Prompt, memory, ranking, and tools \\
    Stochastic sensitivity & 200 & Sampled-output check \\
    \bottomrule
  \end{tabular}
\end{table}

\section{Experiments and Results}

\paragraph{Experimental setup.}
We evaluate Llama 3.1-8B, 4-bit Mistral Small 3.1-24B, and Qwen 2.5 3B/7B/14B using the same prompt, actions, parser, simulator, safety checks, score, and deterministic decoding. Each model runs on 800 primary cases under five settings (4,000 episodes). To estimate uncertainty, we repeatedly resample the 800 physical cases 2,000 times and recompute each metric. The five observation versions of a case are always kept together, so repeated views of the same fault are not treated as independent evidence. Paired diagnostics reuse the same cases and correct for testing several interface changes. Model revisions and full statistical details are in the appendix.

\paragraph{Primary performance.}
Every model has median regret zero, so at least half of its episodes match the oracle reference cost. The mean ranking is instead driven by a smaller failure tail: P95 regret is 0.814 for Qwen 7B and 0.873--1.000 for the others, meaning that their worst episodes approach the cost of taking no action. Qwen 7B has the best mean and tail regret, whereas Qwen 3B has the highest regret and lowest restored fraction. Because Qwen 7B also beats the larger Qwen 14B and Mistral 24B, parameter count alone does not explain performance.

Restored fraction records the final outcome, while regret also accounts for outage duration and customer priority. Thus Qwen 7B and Qwen 14B both restore 94.9\% of customers, but Qwen 7B scores better by restoring important loads sooner. Under noisy observations, idealized FLISR and greedy obtain regret 0.054 and 0.060 versus 0.089 for the best language model; FLARE therefore does not imply that language models replace specialized methods.

\paragraph{Performance across observation and operating conditions.}
\begin{table}[t]
  \centering
  \scriptsize
  \caption{Primary results across five observation settings. Lower regret, P95, and unsafe proposals are better; higher restored fraction is better.}
  \label{tab:primary}
  \setlength{\tabcolsep}{2pt}
  \renewcommand{\arraystretch}{0.95}
  \resizebox{\linewidth}{!}{%
  \begin{tabular}{@{}lrrrr@{}}
    \toprule
    Model & Mean regret [95\% CI] & P95 & Restored & Unsafe/episode \\
    \midrule
    Llama 3.1-8B & 0.124 [0.109, 0.139] & 1.000 & 0.945 & 0.18 \\
    Mistral Small 24B, 4-bit & 0.088 [0.073, 0.103] & 0.969 & 0.944 & 0.16 \\
    Qwen 2.5-3B & 0.148 [0.132, 0.164] & 1.000 & 0.892 & 0.21 \\
    Qwen 2.5-7B & \textbf{0.082 [0.068, 0.097]} & \textbf{0.814} & \textbf{0.949} & 0.17 \\
    Qwen 2.5-14B & 0.111 [0.096, 0.125] & 0.873 & \textbf{0.949} & \textbf{0.16} \\
    \bottomrule
  \end{tabular}}
\end{table}

Clean observations give the lowest regret for every model; the conflicting condition, which combines lower SCADA coverage with higher channel and inspection errors, is hardest. Because these factors vary together, this comparison does not isolate a conflict-only effect. Llama rises from 0.065 (clean) to 0.158 (conflicting), while Qwen 7B remains within 0.084--0.089 across degraded settings. The models therefore respond differently to compound information degradation.

The 15 topology, DER-ratio, and loading contrasts range from $-0.023$ to $+0.017$, and every confidence interval includes zero. We detect no consistent loss under these shifts, but three controlled 200-case panels cannot establish general robustness.

\paragraph{Interface diagnostics and safety.}
Schema-only instructions increase Qwen 3B regret by $+0.668$ [$0.648,0.690$] but decrease Llama 8B regret by $-0.049$ [$-0.056,-0.042$]. The smaller Qwen appears to rely on the procedural explanation, whereas the same detail may distract Llama. Memory, simulation, and ranking changes also have model-dependent signs (Appendix~\ref{app:results}); no component is universally helpful.

Shadow replay finds that 15,142 of 15,706 guard rejections (96.4\%) would cause a physical violation. Thermal overload is most common (8,165), consistent with ties transferring demand to a loaded alternate feeder; labels overlap. Zero realized hard failures therefore demonstrate the guard, not intrinsically safe proposals. Balanced positive-sequence power flow also omits native three-phase and short-circuit validation, so the results support evaluation rather than autonomous operation.

\section{Conclusion and Future Work}

\bench{} establishes a reproducible and physics-grounded foundation for evaluating language-model agents in post-isolation feeder restoration. Our results demonstrate that these agents can make effective restoration decisions across diverse grid and observation conditions while also revealing clear opportunities to improve reliability in difficult cases. This provides encouraging evidence for their future use as decision-support tools for grid operators. Future work will extend \bench{} to the complete FLISR workflow, incorporate unbalanced three-phase and protection models and validate agent assistance using utility data and human-operator studies.

\typeout{NEURIPS-MAIN-CONTENT-END-PAGE=\thepage}

\section*{Acknowledgment}
The authors acknowledge Mahindra University for access to the DGX supercomputing facility and the technical and financial support from the Mahindra University--Virginia Tech Interdisciplinary Advanced Research Centre for Transformative Technologies (MU-VT IARCT$^{2}$), Project No. MU-VT IARCT$^{2}$/TA-AIML/2025/01.

\paragraph*{LLM-use disclosure.}
Language models assisted with coding and manuscript preparation. All ideas, analyses, results, and conclusions belong to the authors, who take full responsibility for the paper.

\bibliographystyle{plainnat}
\bibliography{references}

\clearpage
\appendix
\section{Benchmark Settings}
\label{app:data}

The five settings change only the information shown to the agent. Coverage is the share of buses with SCADA; FP/FN are call false-positive/negative rates; FI is fault indicator. SCADA age is metadata in minutes, not replayed historical state. Flag $p$ is the probability of an internal contradictory-evidence marker.

\begin{table}[h]
  \centering
  \scriptsize
  \caption{Observation settings. Values are probabilities unless stated otherwise.}
  \label{tab:tiers}
  \setlength{\tabcolsep}{3.1pt}
  \begin{tabular}{@{}lrrrrrrr@{}}
    \toprule
    Setting & Coverage & Call FP & Call FN & FI error & Crew error & SCADA age & Flag $p$ \\
    \midrule
    Clean & 1.00 & 0 & 0 & 0 & 0 & 0 min & 0 \\
    Sparse & 0.35 & .05 & .10 & .05 & .02 & 5 min & 0 \\
    Noisy & 0.35 & .20 & .25 & .20 & .05 & 10 min & 0 \\
    Stale & 0.35 & .20 & .25 & .20 & .05 & 45 min & 0 \\
    Conflicting & 0.25 & .30 & .30 & .30 & .08 & 30 min & .35 \\
    \bottomrule
  \end{tabular}
\end{table}

Agents can request reports, inspect a location, read a fault indicator, test or simulate a switch, open a sectionalizer, close a tie, or stop. Table~\ref{tab:budgets} lists the main limits.

\paragraph{Report contents and degradation.}
SCADA returns energized status and voltage for a tier-dependent subset of instrumented buses, with a \texttt{stale\_min} age tag; readings are generated from the current observable state. Trouble-call/AMI reports provide a noisy aggregate count, fault indicators return a Boolean for a queried bus, and the network model exposes as-built topology without live state. Optional crew inspection returns a possibly erroneous fault-present result and costs 45 minutes; switch testing returns only \texttt{held}, \texttt{re\_tripped}, or \texttt{uninformative}. Simulated-switch feedback is a coarse feasibility class computed from report-derived beliefs.

The final table column is an internal marker probability: current tool responses do not forward this marker to the agent, and no report pair is explicitly forced to disagree. Channel-specific errors can nevertheless yield inconsistent readings. Thus the conflicting condition is a compound coverage/error setting, not an isolated conflict intervention.

\begin{table}[h]
  \centering
  \scriptsize
  \caption{Episode limits.}
  \label{tab:budgets}
  \begin{tabular}{@{}lr@{}}
    \toprule
    Quantity & Value \\
    \midrule
    Restoration horizon & 240 minutes \\
    Switching operations & 4 \\
    Sectionalizer openings & 3 \\
    Tool calls & 20 \\
    AC simulations & 8 \\
    Customer load & 1.2 kW per customer \\
    Customer-class weights & 1, 1.5, 2, 3 \\
    \bottomrule
  \end{tabular}
\end{table}

The topology shift changes normal-open points. The generation shift increases the DER-to-load ratio by 50\%. The loading shift moves the operating point toward 95\% equipment loading. Every shifted case is checked again for feasibility.

\section{Evaluation Details}
\label{app:evaluation}

\begin{table}[h]
  \centering
  \scriptsize
  \caption{Models used in the experiments. Revisions are shortened for readability.}
  \label{tab:models}
  \setlength{\tabcolsep}{3pt}
  \begin{tabular}{@{}l l l@{}}
    \toprule
    Model & Revision & Format \\
    \midrule
    Llama 3.1-8B & \texttt{0e9e39f249a1} & Standard \\
    Mistral Small 24B & \texttt{b28a0cfd2f7f} & 4-bit \\
    Qwen 2.5-3B & \texttt{aa8e72537993} & Standard \\
    Qwen 2.5-7B & \texttt{a09a35458c70} & Standard \\
    Qwen 2.5-14B & \texttt{cf98f3b3bbb4} & Standard \\
    \bottomrule
  \end{tabular}
\end{table}

All runs used one NVIDIA A100 40GB GPU. Decoding was deterministic. Mistral used 4-bit weights; the other models used their standard weights.

Reference policies include no action, random feasible actions, greedy restoration, FLISR, a belief planner, and the case-level oracle. The oracle searches feasible controller states under the four-operation and three-opening budgets, using the shared weighted outage cost and serialized operation times (one minute for remote switches; manifest travel time for manual switches). Each restored weighted customer accrues cost until its operation completes; customers still out accrue the full 240-minute horizon. The oracle returns the lowest-cost feasible sequence found and certifies optimality only if the finite search exhausts its states. Candidate closures use the environment's radiality, fault-energization, and AC checks; openings use dark-sectionalizing rules. Agent-facing switch simulations estimate feasibility from reports, while the pre-commit guard uses the simulated physical state.

For confidence intervals, we resample physical cases 2,000 times and recompute the result. The five observation settings for the same case stay together. Diagnostic comparisons use the same cases on both sides. All result files were checked for missing episodes, duplicate cases, invalid scores, and revision mismatches.

\section{Additional Results}
\label{app:results}

\begin{figure}[h]
  \centering
  \includegraphics[width=\linewidth]{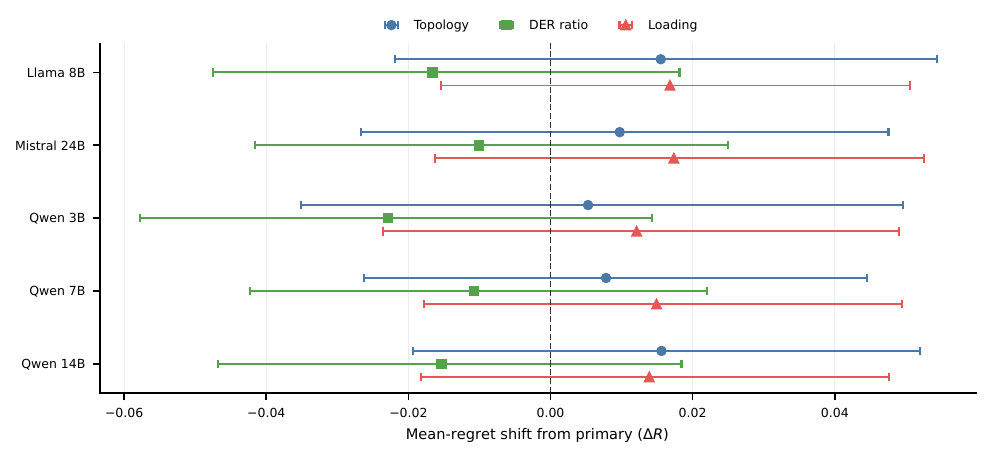}
  \caption{Change in mean regret under topology, DER, and loading shifts. Bars show 95\% confidence intervals. Every interval crosses zero.}
  \label{fig:ood}
\end{figure}

We do not find a consistent effect from the three operating shifts. This does not prove that the models are robust to other grid changes.

\begin{table}[h]
  \centering
  \scriptsize
  \caption{Change in mean regret after each interface change. Positive values are worse.}
  \label{tab:ablations}
  \setlength{\tabcolsep}{2.7pt}
  \begin{tabular}{@{}lrrrrr@{}}
    \toprule
    Model & No memory & No simulator & No ranking & Schema only & True state \\
    \midrule
    Llama 8B & +.046 & $-.021$ & +.066 & $-.049$ & $-.028$ \\
    Mistral 24B & $-.006$ & .000 & $-.015$ & +.172 & $-.002$ \\
    Qwen 3B & +.514 & $-.001$ & +.073 & +.668 & +.045 \\
    Qwen 7B & $-.006$ & +.001 & $-.014$ & $-.007$ & +.003 \\
    Qwen 14B & $-.021$ & $-.007$ & +.070 & +.089 & $-.033$ \\
    \bottomrule
  \end{tabular}
\end{table}

The interface changes do not affect every model in the same way. This is why we report each model separately.

\section{AI4PowerGrids Checklist}
\label{app:checklist}

\paragraph{Physics feasibility.}
Section 2 uses balanced nonlinear AC power flow rather than a DC approximation. Section 3 reports 15,706 guard rejections: 15,142 would be unsafe and 8,165 include thermal overload. No unsafe action was applied.

\paragraph{Out-of-distribution evaluation.}
Table 1 and Figure 1 cover held-out topology, DER-ratio, and loading shifts created after the benchmark was frozen. The evaluated models were not trained or fine-tuned on FLARE cases.

\paragraph{Failure modes.}
Table 2 reports P95 regret, and Section 3 explains the near-no-action failure tail. Section 3 and Appendix C also report guard failures and interface sensitivity.

\paragraph{System-level effect.}
Section 2 defines outage cost using weighted energy not served. Table 2 reports regret, restored customers, and unsafe proposals, linking decisions to service restoration and grid safety.

\paragraph{Data and code.}
Code is available at \url{https://github.com/dr-pandit-69/FLARE-neurips}. Benchmark data and configurations will be released on acceptance. Model weights are referenced but not redistributed.

\end{document}